\pdfoutput=1

\documentclass[11pt]{article}
\usepackage{url}
\usepackage{comment}
\usepackage{pdfpages}
\usepackage {array,booktabs}
\usepackage[]{acl}
\usepackage{here}
\usepackage{times}
\usepackage{latexsym}

\usepackage[T1]{fontenc}

\usepackage[utf8]{inputenc}

\usepackage{microtype}

\title{Probing Physical Reasoning with Counter-Commonsense Context}

\author{Kazushi Kondo,$^1$ Saku Sugawara,$^2$ Akiko Aizawa$^2$ \\
 $^1$The University of Tokyo, $^2$National Institute of Informatics \\
  \texttt{kkkazu@g.ecc.u-tokyo.ac.jp, \{saku,aizawa\}@nii.ac.jp}}

\begin{document}
\maketitle
\begin{abstract}
In this study, we create a CConS (Counter-commonsense Contextual Size comparison) dataset to investigate how physical commonsense affects the contextualized size comparison task; the proposed dataset consists of both contexts that fit physical commonsense and those that do not.
This dataset tests the ability of language models to predict the size relationship between objects under various contexts generated from our curated noun list and templates.
We measure the ability of several masked language models and generative models.
The results show that while large language models can use prepositions such as ``in'' and ``into''  in the provided context to infer size relationships, they fail to use verbs and thus make incorrect judgments led by their prior physical commonsense.
\end{abstract}

\section{Introduction}

Humans possess physical commonsense regarding the behavior of everyday objects.
Physical commonsense knowledge is relevant to their physical properties, affordances, and how they can be manipulated \cite{bisk_piqa_2020}. 
While a significant amount of physical commonsense can be expressed in language \cite{forbes_verb_2017,bisk_piqa_2020}, direct sentences describing facts such as ``people are smaller than houses'' rarely appear because of reporting bias \cite{gordon_reporting_2013,ilievski_dimensions_2021}.
Recent language models have succeeded in tasks that do not require contextual reasoning, such as size comparison and prediction of event frequency \cite{talmor-etal-2020-olmpics}.

However, what about inferences that are context-dependent?
Whether a language model can make correct inferences in various contexts is important because physical reasoning is highly context-dependent \cite{ogborn_science_2011}.
Several studies on contextual physical reasoning \cite{forbes_neural_2019,bisk_piqa_2020,aroca-ouellette_prost_2021,zellers_piglet_2021} have been conducted to produce datasets that assess the ability to recognize physical situations described in writing. Without context, however, these datasets may be answered by commonsense.
\begin{figure}[t]
    \centering
    \includegraphics[width=\linewidth]{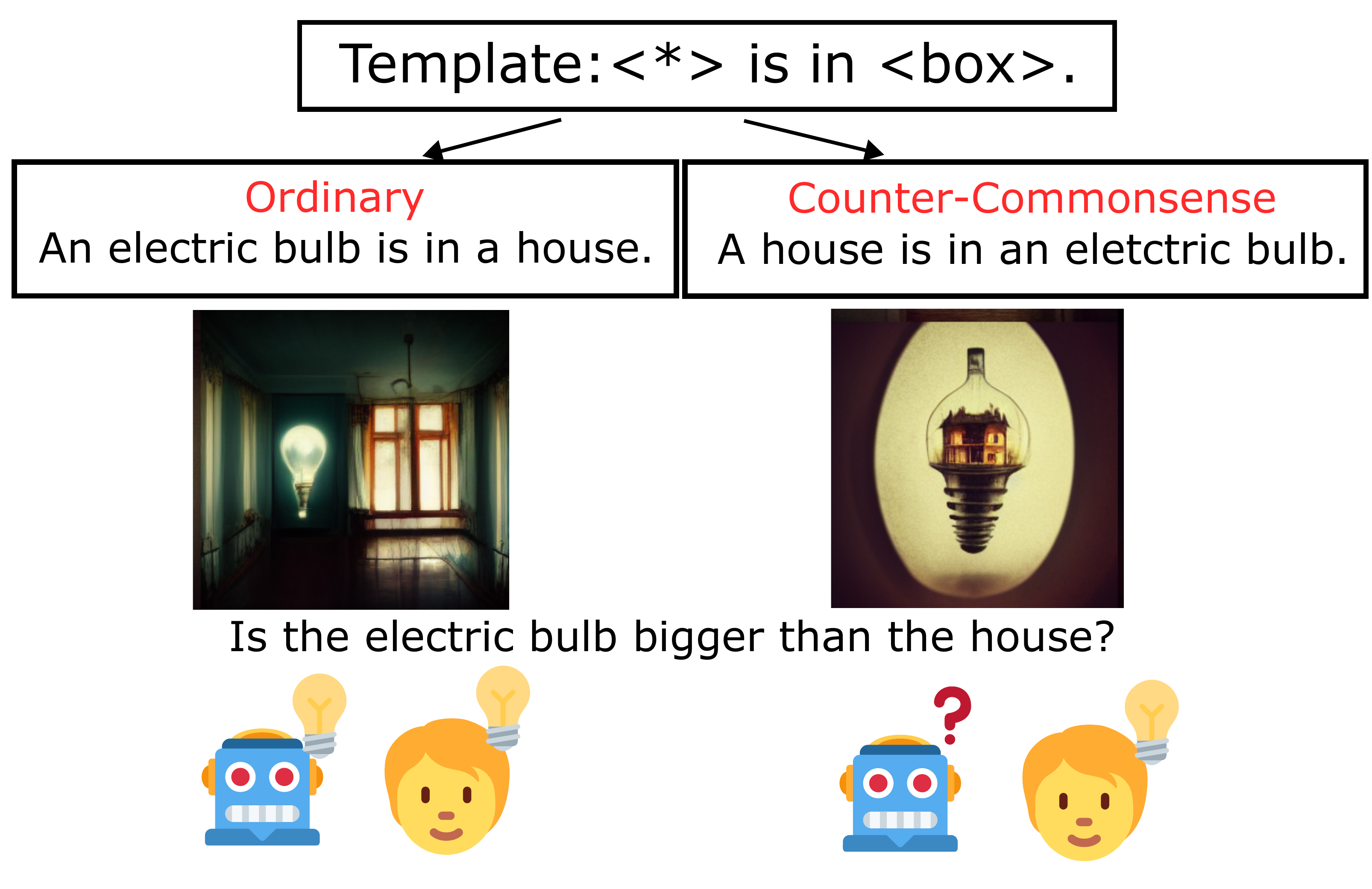}
    \caption{Examples of contexts that do or do not accord with ordinary commonsense.
 Humans can imagine the situation and make correct inferences, but language models are drawn to commonsense and make incorrect judgments.
 The example images are generated by Midjourney (\url{https://midjourney.com
}).}
    \label{fig:title}
\end{figure}

Humans also can reason in ways that differ from simply using commonsense.
For instance, if the context ``there is a house inside a light bulb.'' is provided, humans can still imagine the situation and reason that the bulb must be larger than the house.
In other words, commonsense is just a sweeping generalization, and reasoning about context must be independent of commonsense.
This reasoning with defeasibility, which reflects the ability to reason logically without relying only on commonsense, seems to have been overlooked in the study of language models compared to the acquisition of commonsense.
Previous investigations of contextual physical reasoning \cite{aroca-ouellette_prost_2021,yu_pacs_2022} failed to distinguish physical reasoning from the simple use of physical commonsense.
To appropriately measure physical reasoning ability, we must use contexts that go against commonsense to rule out the possibility that the model is overconfident in physical commonsense.

In this study, we investigate the behavior of the language model concerning physical commonsense given the context of a situation that contradicts commonsense. 
We choose the size comparison task despite various possible domains of physical commonsense \cite{ilievski_dimensions_2021}.
The task is one of the easiest physical commonsense reasoning tasks for language models \cite{forbes_verb_2017,goel_how_2019}, and it is also easy to add a context to change the relationship between sizes.
For example, in this study, the context is a sentence that implies a size relationship, such as ``<obj1> contains <obj2>.'' 

For this purpose, we created a new dataset, CConS (Counter-commonsense Contextual Size comparison)\footnote{\url{https://github.com/cfkazu/Counter-Commonsense-Context}}.
This dataset contains 1,112 sentences generated from 139 templates and tests the ability of language models to infer the size relationship between objects using a cloze-style prompt.
Figure \ref{fig:title} shows the size comparison examples with or without contexts that (do not) agree with ordinary commonsense.
Our experiments using recent language models show that GPT-3(text-davinci-003) \cite{brown_language_2020} correctly reasons in context when it is consistent with commonsense, yielding 85\% accuracy. 
In contrast, even GPT-3 can only show poor performance (41 \% accuracy) for examples that contradict commonsense.
This suggests that the models may not effectively distinguish between physical commonsense and inferences based on contexts, leading to incorrect predictions.
Nevertheless, when prepositions hint at the relationships, the accuracy rate exceeded 55\%, even for counter-commonsense examples.
In summary, our counter-commonsense examples reveal the difference in influence between prepositions and verbs in contextualized physical reasoning. 

The contributions of this study are as follows:
  \begin{enumerate}
    \item We create a dataset that assesses size comparison ability more precisely by contrasting examples that conform to physical commonsense with ones that do not.
    \item We show that physical commonsense prevents measuring the language models' ability of contextual physical reasoning.
    \item We demonstrate that even large models perform poorly when making inferences that violate physical commonsense. 
    Specifically, they struggle to infer size relations implied by verbs and can infer only when prepositions indicate.
\end{enumerate}

\section{Related Works}
\paragraph{Size Comparison Task}
The size comparison task, which previous studies \cite{yang_extracting_2018,goel_how_2019} investigated since the earlier linguistic representations, such as GloVe \cite{pennington_glove_2014} or ELMo \cite{peters-etal-2018-deep}, is one of the easiest physical common-sense inference tasks for language models \cite{forbes_verb_2017,goel_how_2019}. 
While there are many prior studies \cite{elazar_how_2019,zhang_language_2020} on this topic, VerbPhysics \cite{forbes_verb_2017} is the most similar to this study in that it focuses on the relationship between sizes and verbs.
There are also some other approaches, such as methods that extract external knowledge \cite{elazar_how_2019}, filling-masks \cite{talmor-etal-2020-olmpics}, or generate images \cite{liu-etal-2022-things}.
These results suggest that the commonsense of comparing object size is encoded in recent language models.
However, these studies do not consider the context that might influence the results of size comparisons.

\paragraph{Defeasible Reasoning}
According to \citet{koons_defeasible_2022}, defeasible reasoning is an argument that is rationally persuasive but not completely valid as a deduction.
This defeasible reasoning is similar to the subject of this study in that it involves the recognition that commonsense and assumptions in a given context are not entirely correct propositions.
Therefore, this study can be seen as an investigation into whether a language model can capture commonsense as defeasible reasoning. 
The creation of a dataset dealing with defeasible reasoning has been discussed by \citet{rudinger_thinking_2020} and \citet{allaway_penguins_2022}. 
Our study is similar to \citet{allaway_penguins_2022} in that it generates sentences that violate the context by fitting words to a template. However, this study differs in that we also generate examples contrary to commonsense for measuring the actual performance of the language model as well as the differences from the ordinary case.

\section{Dataset Creation}
\begin{table*}
\centering
\small
\begin{tabular}{lll}
\toprule
\textbf{Template}&\textbf{Generated: Ordinary Examples} &\textbf{Generated: Counter-Commonsense Examples}\\
\midrule
He found <portable> in <box>. &He found a key in a key box. &He found a monitor in a key box.  \\
<box> contains <portable>. & A key box contains a key. & A key box contains a monitor.\\
<*> fills <box>. &A marble fills a bin.&A refrigerator fills a bin.\\  
<*> is covered by <flat>. & A pen is covered by a newspaper. & A desk is covered by a handkerchief.\\
\bottomrule
\end{tabular}

\caption{\label{citation-guide}
Examples of the templates.
<tag> constrains possible nouns to be filled.
For example, <box> means that the noun entering there must have the attribute ``box,'' that is, it must be able to hold things.
<*> indicates that any words in the noun list (only material nouns) can be inserted.
}
\label{table_template}
\end{table*}

In this study, we create 139 templates and automatically generate 1,112 examples.
Table \ref{table_template} lists examples of these templates.

\paragraph{Designing Template}
We focus on the comprehensiveness of verb phrases while designing templates to ensure that the choice of verbs is not arbitrary.
Therefore, we extract 139 verb phrases that indicate size relationships from the Oxford 5000 dictionary \footnote{\url{https://www.oxfordlearnersdictionaries.com/about/wordlists/oxford3000-5000}} and manually assemble simple sentences.
For example, the statement ``<obj1> beats <obj2>'' 
is not included in this template because this statement is not informative enough to determine a size relation.

Moreover, in comparing sizes, we also notice not only verbs but the usage of prepositions such as ``in'' or ``into'' may provide clear clues about the size relationships. 
Therefore, we select templates that contain only examples with these prepositions and distinguish them as easy templates from those that do not as hard templates.
In subsequent experiments, we also investigate the effect of this difference on the behavior of the language model.

\paragraph{Restriction on Noun}
 If nouns are arbitrarily inserted, the resulting sentences may be nonsensical or impossible for a human to imagine.
For example, we choose not to include the sentence ``the stone threw the dog'' because it is beyond imagination. 

We place restrictions on the nouns used in the sentence templates by defining tags to avoid this nonsense. 
A single placeholder can have constraints (multiple tags).
There are 18 types of tags, including ``have\_hands,'' ``box,'' and ``portable.''
Tags are manually determined to abstract the properties of verb phrases.
We also use the Oxford 5000 dictionary to obtain a list of nouns referring to physical objects.
One of the nouns that satisfy all constraints is randomly selected from a list of 195 nouns and inserted.
\paragraph{Generating Sentences}
\label{sec:generating}
The template tags are replaced with the corresponding nouns to generate the context, and the questions asking for size comparisons are combined.
For example, the contextualized question text provided to the masked language models is as follows:\\``\texttt {\small<<context>> In this situation, the size of <obj1> is probably much [MASK] than the size of <obj2>.}''

Contexts and questions are used to generate input for each of the masked language models and generative models.
We classify generated sentences to the Ordinary or Counter-Commonsense (CCommon) subset based on whether the size relationship between objects indicated by the template accords commonsense.

\section{Experiment}

\paragraph{Task Definition}
We measure the ability of masked language models and generative models to recognize size relationships by providing sentences for each architecture.
These sentences are generated from templates (Section \ref{sec:generating}).
We also see how the language model's behavior changes when context sentences follow or do not follow a general common-size relationship.

\paragraph{Comparison Aspects}
We investigate how language models create physical reasoning without being biased by their prior physical commonsense.
\begin{enumerate}
    \item How do the physical reasoning results of the language model change when contexts are consistent or inconsistent with commonsense?
    
    \item 
    How does the performance of a language model change when comparing an easy dataset that contains certain prepositions that hint at size relationships with a hard dataset that does not?
\end{enumerate}

\paragraph{Model Settings}
In this study, BERT \cite{devlin_bert_2019}, RoBERTa \cite{liu_roberta_2019}, and ALBERT \cite{lan_albert_2020} are used to assess the performance of the masked language models.
We also investigate how the size of the model affects physical reasoning.
We choose T0 \cite{sanh2022multitask} and GPT-3(text-davinci-003) to evaluate the performance of the generative model.

According to \citet{talmor-etal-2020-olmpics}, RoBERTa-Large outperforms BERTs and RoBERTa-Base in a no-context size comparison task. 
Proceeding from this analysis we attempt to detect whether commonsense influences physical reasoning by giving examples contrary to commonsense as context.


\paragraph{Tasks Format Details}The tasks are performed by inputting sentences according to the format defined for each of the models, as follows.
\subparagraph{Format for Masked Language Models\\}
\textbf{WithContext:\;}\texttt{<<context>> In this situation, the size of <obj1> is probably much [MASK] than the size of <obj2>.}\\
\textbf{WithoutContext:\;}\texttt{The size of <obj1> is probably much [MASK] than the size of <obj2>.}

The candidates for \texttt{[MASK]} are ``larger,'' ``bigger,'' ``smaller,'' and ``shorter.''
If the sum of the probabilities of the first two options exceeds 0.5, language models predict that obj1 is larger than obj2.
Therefore, the language model always makes binary decisions.

\subparagraph{Format for Generative Models\\}
\textbf{WithContext:\;}\texttt{<<context>> Which is bigger in this situation, <obj1> or <obj2>?}\\
\textbf{WithoutContext:\;}\texttt{Which is bigger in general, <obj1> or <obj2>?\\}
\texttt{<<context>>} is a sentence generated from templates.

\paragraph{Human Evaluation}
We ask crowdworkers to perform the same size comparison task to measure the accuracy of humans in this task. 
Thus, we can test the validity of the automatically generated questions.
The crowdworkers are given the same context and make a choice that is larger. (See Appendix \ref{sec:appendix_human} for details.)
Five crowdworkers are assigned to each question. 
We use some intuitive examples, such as ``<obj1> contains <obj2>,'' which are provided for qualification, and exclude those who get such examples wrong or choose the same answer for all examples. 

\begin{table}

\small
\begin{tabular}{llll}
\toprule
\centering
\textbf{Model} & \textbf{Ordinary}& \textbf{CCommon} &\textbf{NoCon} \\
\midrule
BERT-B&0.483&0.515&0.495\\
BERT-L & 0.500&0.521&0.494 \\
RoBERTa-B&0.554&0.443&0.507\\
RoBERTa-L & 0.692&0.413&0.639 \\
ALBERT-B&0.500&0.521&0.494\\
ALBERT-XXL&0.720&0.346&0.701\\
T0++&0.682&\textbf{0.530}&0.589\\
T0&0.684&0.443&0.574\\
GPT-3&\textbf{0.856}&0.415&\textbf{0.764}\\
\midrule
Human&0.814&0.798&0.791\\
\bottomrule
\end{tabular}
\caption{\label{citation-guide}
The inference results of the language model for data sets where the context follows and does not follow commonsense and context is removed.
}
\label{table_auto}
\end{table}
\section{Result and Analysis}

Tables \ref{table_auto} and \ref{table_auto_divided} exhibit the performance of the language model on our datasets.
GPT-3 outperforms other models in Ordinary and NoCon setups.
RoBERTa-Large and ALBERT-XXLarge show better reasoning ability than the other masked language models in the Ordinary dataset. 
However, for the CCommon dataset, the performance of the pre-trained language model decreases, particularly in ALBERT-XXLarge. 
This result suggests that commonsense built into the model hinders its ability to make accurate judgments.
Other models struggle to capture size relationships.
These results without context (NoCon) are generally consistent with the findings of a previous investigation of the no-context size comparison task conducted by \citet{talmor-etal-2020-olmpics}.

In some CCommon examples, BERT performs better than RoBERTa.
This may be because BERT is less equipped with commonsense, allowing it to make simpler judgments without being influenced.

\paragraph{Impact of Prepositions}
Prepositions did not significantly impact the prediction for the masked language models in the Ordinary dataset. 
However, there is a significant difference in the correct response rates in the CCommon dataset.
RoBERTa-Large performs well in easy data, regardless of whether the context defies commonsense.
This result indicates that RoBERTa-Large recognizes the connection between the prepositions and size relationships.
The ALBERT-XXLarge model does not perform well for the CCommon dataset, even if the setting is easy; therefore, we consider that it merely answers according to commonsense rather than making inferences.
In short, context is not useful for ALBERT when the prepositions do not provide direct hints.

GPT-3 uses prepositions more effectively than other models and performs better on the Easy dataset, while the model struggles to answer the CCommon dataset in the hard setting. This result means GPT-3 learns commonsense well but cannot make physical logical inferences.

 \begin{table}
\centering
\small
\begin{tabular}{lllll}
\toprule
&\multicolumn{2}{c}{\textbf{Ordinary}}&\multicolumn{2}{c}{\textbf{CCommon}}\\
\textbf{Model} & \textbf{Easy}& \textbf{Hard} &\textbf{Easy} &\textbf{Hard}\\
\midrule
BERT-B & 0.506&0.471&0.460&0.557\\
BERT-L &0.527& 0.479&0.480&0.553 \\
RoBERTa-B&0.557&0.550&0.473&0.419\\
RoBERTa-L &0.711& 0.671&0.467&0.369 \\
ALBERT-B&0.527&0.479&0.480&\textbf{0.553}\\
ALBERT-XXL&0.744&0.693&0.353&0.346\\
T0++&0.762&0.607&\textbf{0.593}&0.480\\
T0&0.726&0.638&0.473&0.424\\
GPT-3&\textbf{0.940}&\textbf{0.788}&0.567&0.296\\
\midrule
Human&0.835&0.796&0.829&0.769\\
\bottomrule
\end{tabular}
\caption{
Comparison results of reasoning ability of the language model for datasets that follow the commonsense and those that do not.
Sentences with prepositions ``in'' or ``into'' are included in the easy dataset and otherwise in the hard.
}
\label{table_auto_divided}
\end{table}
\section{Conclusion}
We develop a method providing a counter-commonsense context to measure physical reasoning ability. 
Our proposed contextualized physical commonsense inference dataset reveals that current language models can partially predict size relations but do not perform as well as humans in contexts that contradict commonsense.
These judgments are possible to a limited extent in the presence of certain prepositions such as ``in'' and ``into.''
While we focused on size comparison tasks in this study, the importance of context in physical reasoning is not limited to this task.
Increasing the size and scope of the datasets for contextual commonsense inference is necessary to build language models that more closely resemble humans and differentiate between general commonsense and the facts at hand.



\section*{Limitations}
The main limitation of our method is that it requires human effort to increase the variety of templates, which makes it difficult to create large datasets.
Using templates to generate data reduces the time required to create data manually, but the need for human labor remains an obstacle.
To resolve this, the templates themselves need to be generated automatically, although the tags that constrain the nouns also need to be generated automatically, which is a difficult problem.

 \section*{Acknowledgment}
We would like to thank anonymous reviewers for their valuable comments and suggestions. This work was supported by JST PRESTO Grant Number JPMJPR20C4 and JSPS KAKENHI Grant Number 21H03502.
\nocite{}

\bibliography{anthology,custom,custom2}
\bibliographystyle{acl_natbib}

\appendix

\section{Experiment Details}
We used a language model published on hugging face Transformers \cite{wolf-etal-2020-transformers} except GPT-3 under MIT (RoBERTa) or Apache-2.0 (BERT, ALBERT, T0, T0++) license.
For GPT-3, the OpenAI API (text-davinci-003\footnote{\url{https://platform.openai.com/docs/models/gpt-3-5}}) is used.
All of these models are designed to solve downstream natural language tasks.
Table \ref{table_model} lists the paths for accessing the models via hugging face.

We use a GPU Tesla V100-PCIE-32GB.
The total computation time was 1 hour for the masked language models and 2 hours for the generative models.

\begin{table}
\centering
\small
\begin{tabular}{lll}
\toprule

\textbf{Model} & \textbf{Model-FullName}\\
\midrule
BERT-B & bert-base-uncased &\\
BERT-L &bert-large-uncased \\
RoBERTa-B&roberta-base\\
RoBERTa-L &roberta-large \\
ALBERT-B&albert-base-v2\\
ALBERT-XXL&albert-xxlarge-v2\\
T0++&bigscience/T0pp\\
T0&bigscience/T0\\
\bottomrule
\end{tabular}
\caption{\label{caption_manual}
Paths for using the Hugging Face models used in this study. These models were used without modification.
}
\label{table_model}
\end{table}

\section{Human Evaluation Details}
\label{sec:appendix_human}
\begin{figure*}[h]
    \centering
    \includegraphics[width=\linewidth]{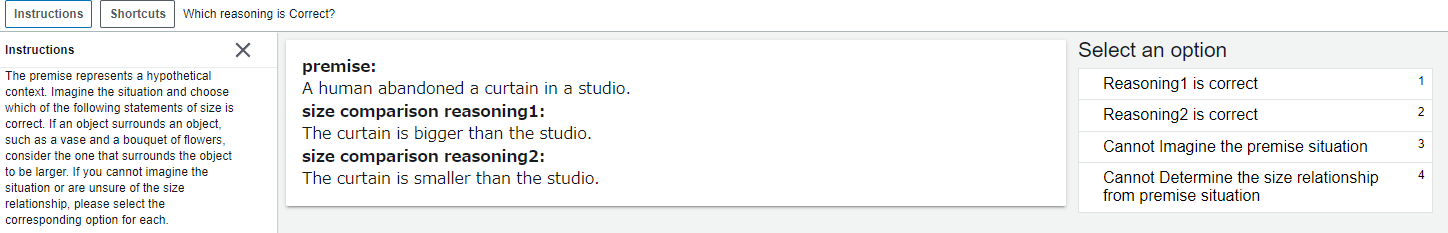}
    \caption{An instruction and options given to Amazon Mechanical Turk crowdworkers for contextualized size comparison task. Annotators are asked to read a context and determine which object is larger in the situation.}
    \label{fig:caption_full}
\end{figure*}

\begin{figure*}[h]
    \centering
    \includegraphics[width=\linewidth]{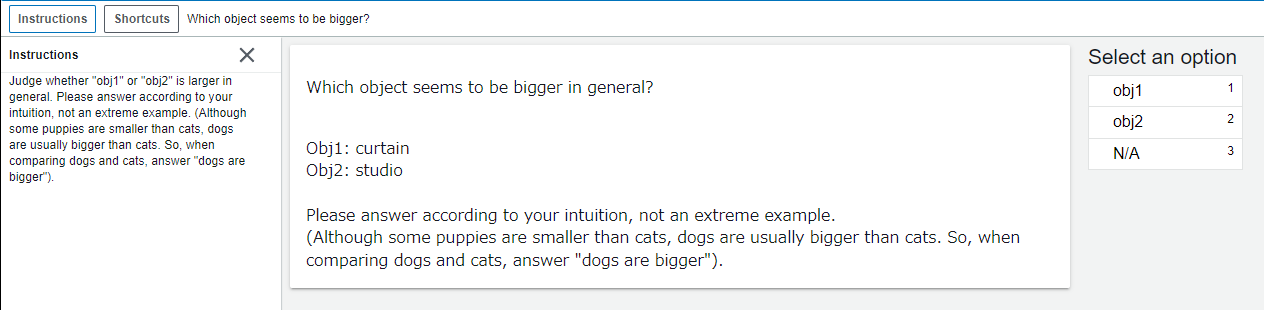}
    \caption{An instruction and options given to Amazon Mechanical Turk crowdworkers for contextualized size comparison task. Annotators are asked to judge which object is generally larger.}
    \label{fig:caption_full_no}
\end{figure*}

We evaluate human accuracy in a size comparison task using Amazon Mechanical Turk. We provide the following instructions and let the crowdworkers choose their answers: We calculate the reward as \$15 per hour.
Figure \ref{fig:caption_full} shows the instructions for the contextualized size comparison task.
The choices are virtually two-option questions, except ``I can't imagine the situation,'' etc.
Figure \ref{fig:caption_full_no} shows the instructions for the non-contextualized size comparison task.
The choices are ``obj1'',``obj2,'' and ``N/A (cannot determine).''

No personal information is obtained. Crowdworkers live in the United Kingdom, the United States, and Canada.
By accepting Amazon Mechanical Turk's participation agreement \footnote{\url{https://www.mturk.com/participation-agreement}}, crowdworkers consent to the collection and use of non-personal data for research purposes.
\end{document}